\documentclass[sigconf,anonymous=false]{acmart}
\usepackage{resizegather}
\usepackage{hyperref}
\usepackage{url}
\usepackage{amsmath}
\usepackage{cleveref}
\usepackage{graphicx}
\usepackage{multirow}
\usepackage{diagbox}
\usepackage{natbib}
\usepackage{libertine}
\usepackage{float}
\usepackage{threeparttable}
\setcitestyle{square,numbers}

\AtBeginDocument{%
  \providecommand\BibTeX{{%
    \normalfont B\kern-0.5em{\scshape i\kern-0.25em b}\kern-0.8em\TeX}}}
\setcopyright{acmcopyright}
\copyrightyear{2020}
\acmYear{2020}
\copyrightyear{2020}
\acmConference{DLP-KDD 2020}{August 24, 2020}{San Diego, California, USA}
\acmBooktitle{San Diego 2020: ACM SIGKDD Conference on Knowledge Discovery and Data Mining,
  August 22--27, 2020, San Diego, CA}
\acmPrice{15.00}
\acmISBN{978-1-4503-XXXX-X/18/06}

\begin{document}

%
% The "title" command has an optional parameter, allowing the author to define a "short title" to be used in page headers.
\title[Research Track Paper]{xDeepInt: a hybrid architecture for modeling the vector-wise and bit-wise feature interactions}

%
% The "author" command and its associated commands are used to define the authors and their affiliations.
% Of note is the shared affiliation of the first two authors, and the "authornote" and "authornotemark" commands
% used to denote shared contribution to the research.
\author{Yachen Yan}
\email{yachen.yan@creditkarma.com}
\orcid{1234-5678-9012}
\affiliation{%
	\institution{Credit Karma}
	\streetaddress{760 Market Street}
	\city{San Francisco}
	\state{California}
	\postcode{94012}
}

\author{Liubo Li}
\email{liubo.li@creditkarma.com}
\orcid{1234-5678-9012}
\affiliation{%
	\institution{Credit Karma}
	\streetaddress{760 Market Street}
	\city{San Francisco}
	\state{California}
	\postcode{94012}
}

\renewcommand{\shortauthors}{Yachen and Liubo}
\renewcommand{\subtitle}{xDeepInt}
%
% By default, the full list of authors will be used in the page headers. Often, this list is too long, and will overlap
% other information printed in the page headers. This command allows the author to define a more concise list
% of authors' names for this purpose.

%
% The abstract is a short summary of the work to be presented in the article.
\begin{abstract}
Learning feature interactions is the key to success for the large-scale CTR prediction and recommendation. In practice, handcrafted feature engineering usually requires exhaustive searching. In order to reduce the high cost of human efforts in feature engineering, researchers propose several deep neural networks (DNN)-based approaches to learn the feature interactions in an end-to-end fashion. However, existing methods either do not learn both vector-wise interactions and bit-wise interactions simultaneously, or fail to combine them in a controllable manner. In this paper, we propose a new model, xDeepInt, based on a novel network architecture called polynomial interaction network (PIN) which learns higher-order vector-wise interactions recursively. By integrating subspace-crossing mechanism, we enable xDeepInt to balance the mixture of vector-wise and bit-wise feature interactions at a bounded order. Based on the network architecture, we customize a combined optimization strategy to conduct feature selection and interaction selection. We implement the proposed model and evaluate the model performance on three real-world datasets. Our experiment results demonstrate the efficacy and effectiveness of xDeepInt over state-of-the-art models. We open-source the TensorFlow implementation of xDeepInt: https://github.com/yanyachen/xDeepInt.

\end{abstract}

\begin{CCSXML}
<ccs2012>
 <concept>
  <concept_id>10010520.10010553.10010562</concept_id>
  <concept_desc>Computing methodologies</concept_desc>
  <concept_significance>500</concept_significance>
 </concept>
 <concept>
  <concept_id>10010520.10010575.10010755</concept_id>
  <concept_desc>Machine learning</concept_desc>
  <concept_significance>300</concept_significance>
 </concept>
 <concept>
  <concept_id>10010520.10010553.10010554</concept_id>
  <concept_desc>Machine learning approaches</concept_desc>
  <concept_significance>100</concept_significance>
 </concept>
 <concept>
  <concept_id>10003033.10003083.10003095</concept_id>
  <concept_desc>Neural networks</concept_desc>
  <concept_significance>100</concept_significance>
 </concept>
</ccs2012>
\end{CCSXML}

\ccsdesc[500]{Computing methodologies}
\ccsdesc[300]{Machine learning}
\ccsdesc{Machine learning approaches}
\ccsdesc[100]{Neural networks}

%
% Keywords. The author(s) should pick words that accurately describe the work being
% presented. Separate the keywords with commas.
\keywords{CTR prediction, Recommendation System, Explicit Feature Interaction, Deep Neural Network}
%
% A "teaser" image appears between the author and affiliation information and the body
% of the document, and typically spans the page.
\maketitle

\section{Introduction}

Click-through rate (CTR) prediction model~\cite{richardson2007predicting} is an essential component for the large-scale recommendation system, online advertising and search ranking~\cite{mcmahan2013ad,he2014practical,cheng2016wide,zhang2019deep}. In online marketplace scenario, accurately estimating CTR will enable the recommendation system to show users the items they prefer to view and explore, which has a direct impact on both short-term revenue and long-term user experience.

The input features (e.g., user id, item id, item category, site domain) of CTR prediction model are usually in a multi-field categorical format~\cite{zhang2016deep} and transformed via field-aware one-hot encoding and multi-hot encoding~\cite{zhou2018deep}. The representation of each field is a sparse binary vector. The corresponding cardinality of each field determines the dimension of the sparse vector. The concatenation of these sparse vectors naturally generates high-dimensional and sparse feature representations.

In CTR prediction model, exploring useful feature interactions plays a crucial role in improving model performance\cite{rendle2010factorization,cheng2016wide,wang2017deep,guo2017deepfm,lian2018xdeepfm}. Traditionally, data scientists search and build hand-crafted feature interactions to enhance model performance based on domain knowledge. In practice, feature interactions of high-quality require expensive cost of time and human workload~\cite{he2014practical}. Furthermore, it is infeasible to manually extract all possible feature interactions given a large number of features and high cardinality~\cite{cheng2016wide}. Therefore, learning low-order and high-order feature interactions automatically and efficiently in a high-dimensional and sparse feature space becomes an essential problem for improving CTR prediction model performance, in both academic and industrial communities.

Deep learning models have achieved great success in recommender systems due to its great feature learning ability. Several deep learning architecture has been proposed from both academia and industry (e.g.,\cite{cheng2016wide,shan2016deep,wang2017deep,qu2016product,qu2018product,guo2017deepfm,lian2018xdeepfm,song2018autoint,huang2019fibinet}). However, All the existing models utilize DNNs as building block for learning high-order implicit bit-wise feature interactions, without bounded order. When modeling explicit feature interactions, the exiting approaches only capture lower order explicit interactions efficiently. Learning higher order typically requires higher computational cost.

In this paper, we propose a efficient neural network-based model called xDeepInt to learn the combination of vector-wise and bit-wise multiplicative feature interactions explicitly. Motivated by polynomial regression, we design a novel Polynomial Interaction Network layers to capture bounded degree vector-wise interactions explicitly. In order to learn the bit-wise and vector-wise interactions simultaneously in a controllable manner, we combine PIN with a subspace-crossing mechanism, which gives a significant boost to our model performance and brings more flexibility. The degree of bit-wise interactions grows with the number of subspace. In summary, we make the following contributions in this paper:
\begin{itemize}
	\item We design a novel neural network architecture named xDeepInt that models the vector-wise interactions and bit-wise interactions explicitly and simultaneously, dispensing with jointly-trained DNN and nonlinear activation functions. The proposed model is lightweight. But it yields superior performance than many existing models with more complex structure. 
	\item Motivated by higher-order polynomial logistic regression, we design a Polynomial-Interaction-Network (PIN) layer which learns higher-order explicit feature interactions recursively. The degrees of interactions are controlled by tuning the number of PIN layers. An analysis is conducted to demonstrate the polynomial approximation properties of PIN.
	\item We introduce a subspace-crossing mechanism for modeling bit-wise interactions across different fields inside PIN layer. The combination of PIN layer and the subspace-crossing mechanism allows us to control the the degree of bit-wise interactions. As the number of subspaces increases, our model can dynamically learn more fine-grained bit-wise feature interactions.
	\item We design an optimization strategy which is in harmony with the architecture of the proposed model. We apply Group Lasso FTRL to the embedding table, which shrinks the entire rows to zero and achieves the feature selection. To optimize weights in PIN layers, we apply FTRL directly. The sparsity in weights results in selection of feature interactions.
	\item We conduct a comprehensive experiment on three real-world datasets. The results demonstrate that xDeepInt outperforms existing state-of-the-art models under extreme high-dimensional and sparse settings. We also conduct a sensitivity analysis on hyper-parameter settings of xDeepInt and ablation study on integration of DNN.
\end{itemize}

\section{Related Work}

Deep learning based models have been applied for CTR prediction problem in the industry since deep neural networks become dominant in learning the useful feature representation of the mixed-type input data and fitting model in an end-to-end fashion\cite{zhang2019deep}. This merit can reduce efforts in hand-crafted feature design and automatically learn the feature interactions.

\subsection{Modeling Implicit Interaction}

Most of the DNN-based methods map the high-dimensional sparse categorical features and continuous features onto a low dimensional latent space in the initial step. Without designing specific model architecture, DNN-based method learns the high-order implicit feature interactions by feeding the stacked embedded feature vectors into a deep feed-forward neural network.

Deep Crossing Network~\cite{shan2016deep} utilizes residual layers in the feed-forward structure to learn higher-order interactions with improved stability. Some hybrid network architectures, including Wide \& Deep Network (WDL)~\cite{cheng2016wide}, Product-based Neural Network (PNN)~\cite{qu2016product,qu2018product}, Deep \& Cross Network (DCN)~\cite{wang2017deep}, Deep Factorization Machine (DeepFM)~\cite{guo2017deepfm} and eXtreme Deep Factorization Machine (xDeepFM)~\cite{lian2018xdeepfm} employ feed-forward neural network as their deep component to learn higher-order implicit interactions. The complement of the implicit higher-order interaction improves the performance of the network that only models the explicit interactions~\cite{beutel2018latent}.

However, this type of approach detects all feature interactions at the bit-wise level~\cite{lian2018xdeepfm} implicitly, without efficiency. And the degree of the interactions are not bounded.

\subsection{Modeling Explicit Interaction}

Deep \& Cross Network (DCN) ~\cite{wang2017deep} explores the feature interactions at the bit-wise level in an explicit fashion. Specifically, each cross layer of DCN constructs all cross terms to exploits the bit-wise interactions. The number of recursive cross layers controls the degree of bit-wise feature interactions.

Some recent models explicitly learn the vector-wise feature interactions using a specific form of the vector product. Deep Factorization Machine (DeepFM)~\cite{guo2017deepfm} combines factorization machine layer and feed-forward neural network through joint learning feature embedding. Factorization machine layer models the pairwise vector-wise interaction between feature $i$ and feature $j$ by the inner product of $\langle x_i, x_j \rangle = \sum_{t=1}^k x_{it}x_{jt}$. Then, the vector-wise interactions are concatenated with the output units of the feed-forward neural network. Product Neural Network (PNN)~\cite{qu2016product,qu2018product} introduces the inner product layer and the outer product layer to learn explicit vector-wise interactions and bit-wise interactions respectively. xDeepFM~\cite{lian2018xdeepfm} learns the explicit vector-wise interaction by using Compressed Interaction Network (CIN) which has an RNN-like architecture and learns all possible vector-wise interactions using Hadamard product. The convolutional filters and the pooling mechanism are used to extract information. FiBiNET~\cite{huang2019fibinet} utilizes Squeeze-and-Excitation network to dynamically learn the importance of features and models the feature interactions via bilinear function.

In the recent research of the sequencing model, the architecture of the Transformer~\cite{vaswani2017attention} has been widely used to understand the associations between relevant features. With different layers of the multi-head self-attentive neural networks, AutoInt~\cite{song2018autoint} can learn different orders of feature combinations of input features. Residual connections~\cite{he2016deep,veit2016residual} are added to carry through different degrees of feature interaction.

The aforementioned approaches learn explicit feature interactions by using outer product, kernel product or multi-head self-attention, which require expensive computational cost.

\section{Model}

\begin{figure}[t]
	\centering
	\includegraphics[width=0.4\textwidth]{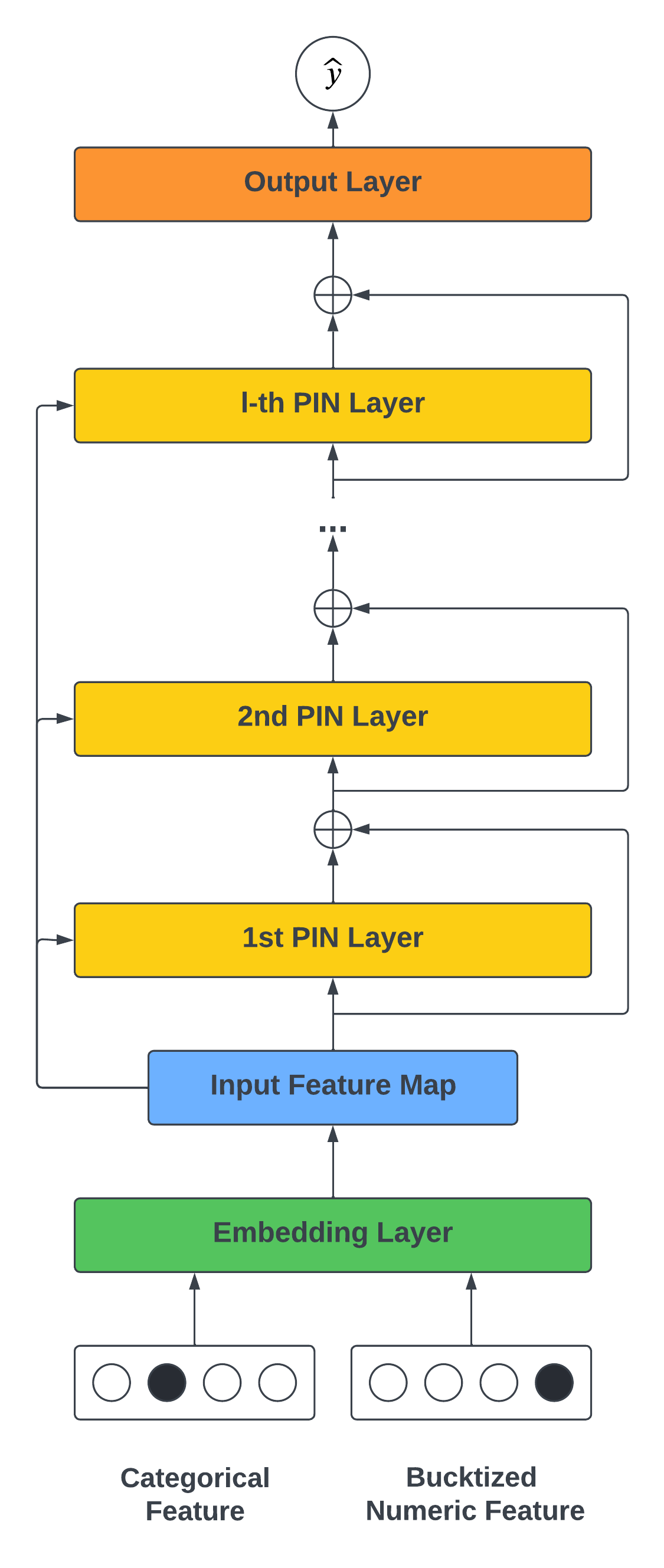}
	\caption{The architecture of unrolled Polynomial Interaction Network with residual connections}
	\medskip
    \small
\end{figure}

In this section, we give an overview of the architecture of xDeepInt. First, we introduce input and embedding layer, which map continuous features and high-dimensional categorical features onto a dense vector. Second, we present the Polynomial Interaction Network(PIN) which utilizes iterative interaction layers with residual connections~\cite{he2016deep,veit2016residual} to explicitly learn the vector-wise interactions. Third, we implement the subspace-crossing mechanism to model bit-wise interaction. The number subspaces controls the degree of mixture of bit-wise and vector-wise interactions.

\subsection{Embedding Layer}

In large-scale recommendation system, inputs include both continuous features and categorical features. Categorical features are often directly encoded by one-hot encoding, which results in an excessively high-dimensional and sparse feature space. 
Suppose we have $F$ fields. In our feature preprocessing step, we bucketize all the continuous features to equal frequency bins, then embed the bucketized continuous features and categorical features to same latent space $R^K$,
\begin{equation*}
	\mathbf{x}^{e}_{f} = \mathbf{x}^{o}_{f}\mathbf{V}_{f},
\end{equation*}
where $\mathbf{x}^{e}_{f}=[x_{f,1},x_{f,2},\cdots,x_{f,K}]$, $x_{f,k}$ is the $k$-th bit of the $f$-th field of the embedding feature map, $\mathbf{V}_f$ is an embedding matrix for field $f$, and $\mathbf{x}^{o}_f$ is a one-hot vector. Lastly, we stack $F$ embedding vectors and obtain an $F$-by-$K$ input feature map $X_0$:
\begin{equation*}
	X_0=\begin{bmatrix}
	    \mathbf{x}^{e}_{1} \\
	    \mathbf{x}^{e}_{2} \\
		\vdots \\
		\mathbf{x}^{e}_{F}
	\end{bmatrix}
\end{equation*}

\subsection{Polynomial Interaction Network}
Consider a $l$-th order polynomial with $f$ variables of the following form:
\begin{equation} \label{equ:poly}
\begin{aligned}
    \Pi_{j=1}^{l}(\sum_{i=1}^f a_{ij}x_{i}+b_{j}).
\end{aligned}
\end{equation}
This polynomial contains all possible multiplicative combinations of $x_i$'s with order less than or equal to $l$ and has an iterative form:
\begin{equation} \label{equ:poly_iterative}
\begin{aligned}
    F^{(l-1)}(x_1,\cdots, x_f)(\sum_{i=1}^f a_{il}x_{i}+b_{l})
\end{aligned}
\end{equation}
where $F^{(l-1)}(x_1,\cdots, x_f)=\Pi_{j=1}^{l-1}(\sum_{i=1}^f a_{ij}x_{i}+b_{j})$. Motivated by the iterative form, we propose polynomial interaction network defined by the following formula:
\begin{equation}\label{interaction-layer}
\begin{aligned}
    X_l & = f(W_{l-1}, X_{l-1}, X_0) + X_{l-1} \\
    & = X_{l-1} \circ (W_{l-1}X_0) + X_{l-1} \\
    & = X_{l-1} \circ [W_{l-1}X_0 + \mathbf{1}] \\
\end{aligned}
\end{equation}
where $\circ$ denotes the Hadamard product. For instance, $[a_{i,j}]_{m\times n}\circ [b_{i,j}]_{m\times n} = [a_{i,j}b_{i,j}]_{m\times n}$. $W_{l-1}\in R^{F\times F}$ and $\mathbf{1}\in R^{F\times K}$ with all entries are equal to one. $X_{l-1}, X_l \in R^{F\times K}$ are the output matrices of ($l$-1)-th and $l$-th interaction layer. Like \eqref{equ:poly}, the $l$-th PIN layer's output is the weighted sum of all vector-wise feature interactions of order less than or equal to $l$.

The architecture of the polynomial interaction network is motivated by the following aspects.

First, the polynomial interaction network has a recursive structure. The outputs of the current layer are built upon the previous layer's outputs and the first order feature map, ensuring that higher-order feature interactions are based on lower-order feature interactions from previous layers.

Second, we use the Hadamard product to model the explicit vector-wise interaction, which brings us more flexibility in crossing the bits of each dimension in shared latent space and preserves more information of each degree of feature interactions.

Third, we build a field aggregation layer $Agg^{(l)}(X)=W_{l}X$ which combines the feature map at the vector-wise level using a linear transformation $W_{l}$. Each vector of the field aggregation feature map can be viewed as a combinatorial feature vector constructed by the weighted sum of the input feature map. Then we take the Hadamard product of the output of the previous layer and field aggregation feature map for this layer. This operation allows us to explore all possible $l$-th order polynomial feature interactions based on existing ($l$-1)-th order feature interactions.

Last, we utilize residual connections~\cite{he2016deep,veit2016residual} in the polynomial interaction network, allowing a different degree of vector-wise polynomial feature interactions to be combined, including the first feature map. Since the polynomial interaction layer's outputs contain all degree of feature interactions, the skipped connection enable next polynomial interaction layer to focus on searching useful higher-order feature interactions while complementing lower-order feature interactions.
\begin{figure}[t]
	\centering
	\includegraphics[width=1\linewidth]{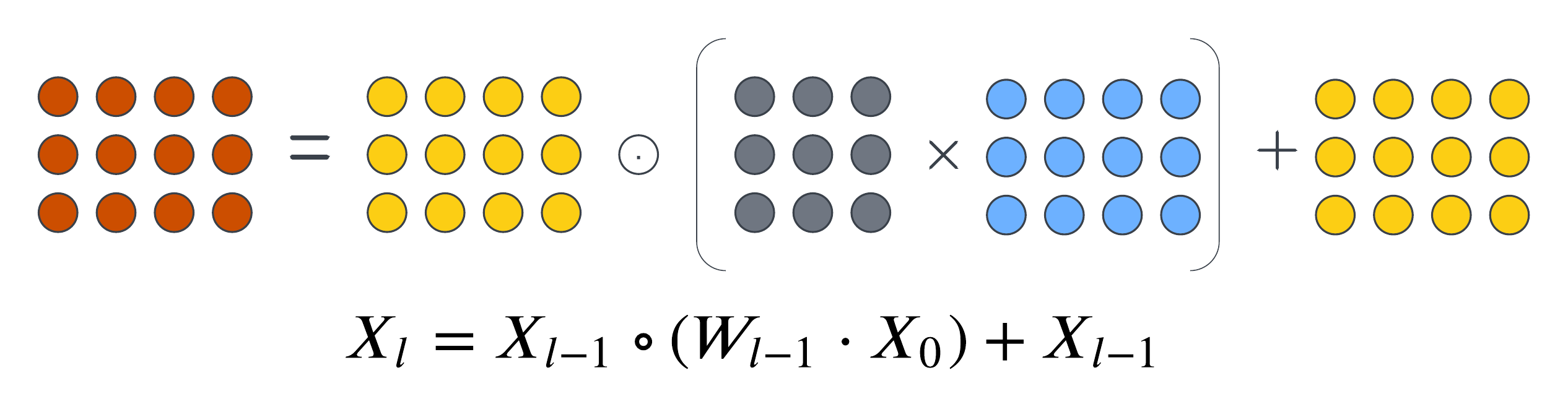}
	\caption{Details of PIN layer}
	\medskip
    \small
    The aggregation layer is defined by $Agg^{(l-1)}(X_0) = W_{l-1} \cdot X_0$. The PIN takes the Hadamard product of the aggregated 1st-order feature map and the output of the previous PIN layer to generate the higher order vector-wise interaction.
\end{figure}
As the number of layer increases, the degree of the polynomial feature interactions increases. The recurrent architecture of the proposed polynomial interaction network enables to bound the degree of polynomial feature interactions.

\subsection{Subspace-crossing Mechanism}

The Polynomial Interaction Network (PIN) models the vector-wise interactions. However, PIN does not learn the bit-wise interaction in the shared latent embedding space. In order to cross the bits of different embedding dimensions, we propose the subspace-crossing mechanism which allows xDeepInt to learn the bit-wise interactions. Suppose we split the embedding space into $h$ sub-spaces, the input feature map $X_{0}$ is then represented by $h$ sub-matrices as follow:
\begin{equation}
\begin{aligned}
    X_0  &  = [X_{0,1}, X_{0,2}, \cdots, X_{0, h}] \\
\end{aligned}
\end{equation}
where $X_{0,i}\in R^{F\times K/ h}$ and $i=1,2\cdots,h$. Next, we stack all sub-matrices at the field dimension and construct a stacked input feature map $X_{0}^{'}\in R^{(F*h)\times (K/h)}$.
\begin{equation}\label{stacked_feature_map}
\begin{aligned}
    X_{0}^{'} & = \begin{bmatrix}
    X_{0,1} \\
    X_{0,2} \\
    \vdots \\
    X_{0,h}
\end{bmatrix},
\end{aligned}
\end{equation}
where $X_{0,j}\in R^{F\times (K/h)}$ and $h$ denotes the number of sub-spaces. By splitting the embedding vector of each field to $h$ sub-vectors and stacking them together, we can align bits of different embedding dimension and create the vector-wise interactions on stacked sub-embeddings.
\begin{figure}[t]
	\centering
	\includegraphics[width=1.0\linewidth]{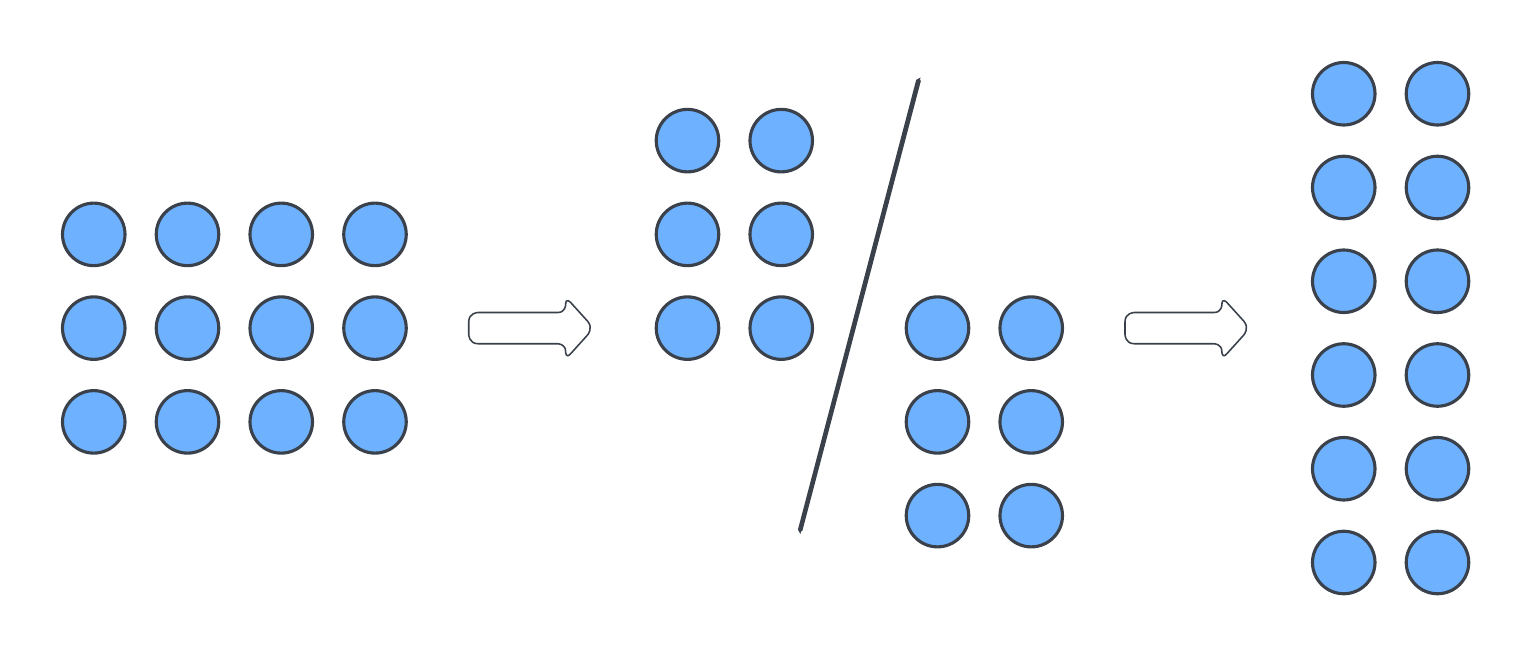}
	\caption{Subspace-crossing mechanism}
\end{figure}
Accordingly, we feed $X_{0}^{'}$ into the Polynomial Interaction Network (PIN):
\begin{equation}
\begin{aligned}
    X_1^{'}  & = X_0
    ^{'}\circ [W_{0}^{'}X_0^{'} + \mathbf{1}] \\
    & \vdots\\
    X_l^{'}  & = X_{l-1}^{'} \circ [W_{l-1}^{'}X_0^{'}+ \mathbf{1}] \\
\end{aligned}
\end{equation}
where $W_{l}^{'} \in R^{(F*h)\times (F*h)}$, $\mathbf{1}\in R^{(F*h)\times (K/h)}$ and $X_{l}^{'} \in R^{(F*h)\times (K/h)}$. 

The field aggregation of feature map and the multiplicative interactions building by Hadamard product are both of the vector-wise level in vanilla PIN layers. The subspace-crossing mechanism enhanced PIN takes the $h$ aligned subspaces as input, so that encouraging the PIN to capture the explicit bit-wise interaction by crossing features of the difference subspaces. The number of subspaces $h$ controls the complexity of bit-wise interactions. Larger $h$ helps the model to learn more complex feature interactions.

\subsection{Output Layer}
The output of Polynomial Interaction Network is a feature map that consists of different degree of feature interactions, including raw input feature map reserved by residual connections and higher-order feature interactions learned by PIN. For the final prediction, we merely use formula as follows:
\begin{equation}
\begin{aligned}
    \hat{y} & = \sigma\big((W_{out}X_{l} + b\mathbf{1}^T)\mathbf{1} \big)
\end{aligned}
\end{equation}
where $\sigma$ is the sigmoid function, $W_{out}\in R^{1\times F}$ is a feature map aggregation vector that linearly combines all the features in the feature map, $\mathbf{1}\in R^K$ and $b\in R$ is the bias.

\subsection{Optimization and Regularization}
For optimization, we use Group Lasso Follow The Regularized Leader (G-FTRL)~\cite{ni2019feature} as the optimizer for the embedding layers for feature selection, and Follow The Regularized Leader (FTRL)~\cite{mcmahan2013ad} as the optimizer for the PIN layers for interaction selection.

Group lasso FTRL regularizes the entire embedding vector of insignificant features in each field to exactly zero, which essentially conducts feature selection and brings more training efficiency for industrial settings. The group lasso regularization is applied prior to the subspace splitting mechanism such that feature selection is consistent between each subspaces.

FTRL regularizes the single elements of weight kernel in PIN layers to exactly zero, which excludes insignificant feature interactions and regularizes the complexity of the model.

This optimization strategy takes advantages of the properties of the different optimizers and achieves row-wise sparsity and element-wise sparsity at embedding table and weight kernel respectively. Therefore, it improves generalization ability and efficiency for both training and serving. It also plays an important role in model compression.

\begin{figure}[t]
\centering
\includegraphics[width=0.4\textwidth]{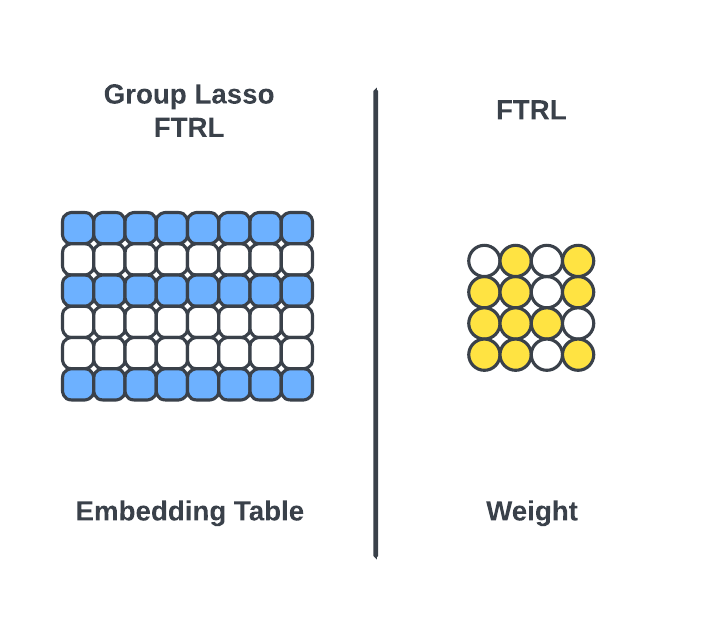}
\caption{Group Lasso FTRL v.s. FTRL}
\medskip
\small
The Group Lasso FTRL regularizes the embedding table with group-wise sparsity. FTRL regularizes the weight kernels of PIN layer with element-wise sparsity.
\end{figure}

\subsection{Training}
The loss function we use is Log Loss,
\begin{align}
    \resizebox{.875\hsize}{!}{$\mbox{LogLoss}=
        -\dfrac{1}{N}\sum^{N}_{i=1}(y_{i}log(\hat{y}_{i})+(1-y_{i})log(1-\hat{y}_{i}))$}
\end{align}
where $y_i$ is the true label and $\hat{y}_{i}$ is the estimated click through rate. $N$ is the total number of training examples.

\subsection{Difference Between PIN and DCN}
PIN and DCN both have an iterative form. However, the two network architectures are quite different in extracting feature interactions.
\begin{equation}\label{DCN}
\begin{aligned}
    x_l & = f_{DCN}(x_{l-1}, w_{l-1}, b_{l-1}) \\
    & = x_0x_{l-1}^Tw_{l-1} + b_{l-1} + x_{l-1}
\end{aligned}
\end{equation}
For DCN, feature map are flattened and concatenated as a single vector. All higher order bit-wise interactions are firstly constructed by the term $x_0x_{l-1}^T$, and then aggregated by a linear regression for next layer. This structure results in a special format of the output. As discussed in \cite{lian2018xdeepfm}, the output the DCN layer is a scalar multiple of $x_0$. \cite{lian2018xdeepfm} also pointed out the downsides: 1) the output of DCN is in a special form, with each hidden layer is a scalar multiple of $x_0$ and thus limits expressive power; 2) interactions only come in a bit-wise fashion.

PIN constructs vector-wise feature interaction using Hadamard product, which preserve the information at vector-wise level. In order to allow different fields to cross at vector level, PIN firstly aggregates the input feature map by a linear transformation $W_{l-1}X_0$ for each iterative PIN layer and build interactions by term $X_{l-1} \circ W_{l-1}X_0$. Accordingly, PIN keeps the vector-wise structure of feature interactions and does not limit the output to a scalar multiple of $X_0$. Moreover, each PIN layer is directly connected with input feature map $X_0$, which improves model trainability. We also prove PIN's polynomial approximation property in later section.

\subsection{xDeepInt Analysis}

In this section, we analyze polynomial approximation property of the proposed xDeepInt model. We consider an xDeepInt model with $l$ PIN layers, a subspace crossing mechanism with $h$ subspaces and $F$ input feature with the same embedding size $K$.
\subsubsection{Polynomial Approximation}
In order to understand how PIN exploits the vector-wise interactions, we examine the polynomial approximation properties of PIN. Let $\mathbf{X}^{(0)} \in R^{F \times K}$ be the embedded feature vector with $\mathbf{x}_i = [x_{i1},\cdots,x_{iK}]$ being the $i$-th row. $\mathbf{x}_{i}^{(l)}=[x_{i1}^{(l)},\cdots,x_{iK}^{(l)}]$ is the $i$-th row of the output of $l$-th layer. Then, $x_{ik}^{(l)}$ has the following explicit form:
\begin{align}
	x_{ik}^{(l)}= x_{ik}^{(0)} \prod_{r=0}^{l-1}(\sum_{i=1}^Fw^{(r)}_{ij}x_{jk}^{(0)}+1), \text{for } k=1,\cdots, K
\end{align}
where $W^{(r)}=[w^{(r)}_{ij}]_{1\leq i,j \leq F}$ is the weight matrix at $r$-th PIN layer. The product $\prod_{r=0}^{l-1}(\sum_{i=1}^Fw^{(r)}_{ij}x_{jk}^{(0)}+1)$ is the weighted sum of all possible crossed terms of the embbeded input at the $k$-th bit having order less than or equal to $l-1$. Thus, $x_{ik}^{(l)}$ is the weighted sum of all crossed terms that contains $x_{ik}^{(0)}$ and has the order less than or equal to $l$.

For the bit-wise interaction modeled by subspace-crossing mechanism, we consider the case where the number of subspaces equals to the embedding size $K$. In this extreme case, each row of $W_0'X_0'$ is a weighted sum of all bits in all fields. This design allows the combination of embedded features at different bits. To be more explicit, we consider the stacked input feature map with $h=K$
\begin{equation}
\begin{aligned}
	X_{0}^{'} & =
	\begin{bmatrix}
		X_{0,1} \\
		X_{0,2} \\
		\vdots  \\
		X_{0,K}
	\end{bmatrix},
\end{aligned}
\end{equation}
where $X_{0,i}=\begin{bmatrix}
x^{(0)}_{i,1} \\
x^{(0)}_{i,2} \\
\vdots \\
x^{(0)}_{i,F}
\end{bmatrix}\in R^F$ .The weight matrix $W'_0$ is given by
\begin{equation}
\begin{aligned}
	W'_0=
	\begin{bmatrix}
		w^{(0)}_{1,1} & w^{(0)}_{1,2} & \cdots &  w^{(0)}_{1,K} & w^{(0)}_{1, K+1} & \cdots & w^{(0)}_{1, FK}\\
		\vdots & \vdots & \ddots & \vdots & \vdots &  \ddots & \vdots \\
		w^{(0)}_{FK,1} & w^{(0)}_{FK,2} & \cdots &  w^{(0)}_{FK,K} & w^{(0)}_{FK, K+1} & \cdots & w^{(0)}_{FK, FK}
	\end{bmatrix} \in R^{FK \times FK}.
\end{aligned}
\end{equation}
Thus, the $k$-th row of $W'_0X'_0 + 1$ has the following form:
\begin{equation}
	\sum_{i=1}^F\sum_{j=1}^Kw^{(0)}_{k,(i-1)K+j}x_{i,j} + 1.
\end{equation}
This is a linear combination of bits in all fields, which allows the PIN exploit all crossing of the feature map at bit-wise level. For example, the $[(i-1)K+j]$-th row of $X'_l$ is given by
\begin{equation}
	x^{(l)}_{i,j}= x^{(0)}_{i,j}\prod_{r=0}^{l-1}\sum_{i=1}^F\sum_{j=1}^Kw^{(r)}_{k,(i-1)K+j}x_{i,j} + 1
\end{equation}
$x^{(l)}_{i,j}$ is the weighted sum of all crossed terms that contains $x^{(0)}_{i,j}$ and has the order less than or equal to $l$.

\subsubsection{Time Complexity}
The cost of computing feature maps $W_{l-1}X$ at $l$-th PIN layer is $\mathcal{O}(h F^2 K)$. For a $L$-layers xDeepInt model, the total cost of feature maps is $\mathcal{O}(L h F^2 K)$. The additional cost is from the Hadamard product and residual connection, which is $\mathcal{O}(L F K)$. In practice, $h$ is not too large. Hence, the total time complexity mainly relies on the number of fields $F$ and embedding size $K$. For an $L$ layers DNN with each layer has $D_k$ hidden nodes, the time complexity is $\mathcal{O}(FK \times D_1 \times D_2 + \sum_{k=2}^{L-1} D_{k-1}D_kD_{k+1})$. The time complexity of xDeepInt relies on the number of subspaces. Therefore, xDeepInt has higher time complexity than DNN when modelling higher degrees of bit-wise interactions.

\subsubsection{Space Complexity}
The embedding layer contains $\sum_{f=1}^F K \times C_f$ parameters, where $C_f$ is the cardinality of $f$-th field. The output layer aggregates the feature map at the last PIN layer. Hence, the output layers require $F + 1$ parameters. The subspace crossing mechanism needs $h^2\times F^2$ parameters at each PIN layer, which is the exact size of the weight matrix $W'_r$ with $0\leq r \leq l-1$.  There are $(K/h) \times k' + h^2\times F^2 \times l$ parameters in $l$ PIN layers. Usually, we will pick a small $h$ to control the model complexity and $k'$ is comparable to $K$. Accordingly, the overall complexity of the xDeepInt is approximate $O(h^2\times F^2\times l)$, which is affected by the embedding size $K$ heavily. A plain $L$ layers DNN with each layer has $D_k$ hidden nodes requires $FK \times D_1 + \sum_{k=2}^{L} D_{k-1}D_{k}$ parameters. The complexity mainly depends on the embedding size and the number of hidden nodes at each layer. To reduce the space complexity of xDeepInt, we can apply the method introduced in~\cite{lian2018xdeepfm}. The space complexity of the model can be further reduced by exploiting a $L$-order decomposition and replace the weight matrix $W'_r$ with two low-rank matrices.

\section{EXPERIMENTS}
In this section, we focus on evaluating the effectiveness of our proposed models and answering the following questions:
\begin{itemize}
	\item \textbf{Q1}: How does our proposed xDeepInt perform in CTR prediction problem? Is it effective and efficient under extreme high-dimensional and sparse data settings?
	\item \textbf{Q2}: How do different hyper-parameter settings influence the performance of xDeepInt?
	\item \textbf{Q3}: Will modeling implicit higher-order feature interactions further improve the performance of xDeepInt?
\end{itemize}

\subsection{Experiment Setup}

\subsubsection{Datasets}
We evaluate our proposed model on three public real-world datasets widely used for research.

\textbf{1. Avazu.}\footnote{https://www.kaggle.com/c/avazu-ctr-prediction} Avazu dataset is from kaggle competition in 2015. Avazu provided 10 days of click-through data. We use 21 features in total for modeling. All the features in this dataset are categorical features.

\textbf{2. Criteo.}\footnote{https://www.kaggle.com/c/criteo-display-ad-challenge} Criteo dataset is from Kaggle competition in 2014. Criteo AI Lab officially released this dataset after, for academic use. This dataset contains 13 numerical features and 26 categorical features. We discretize all the numerical features to integers by transformation function $\lfloor Log\left(V^{2}\right) \rfloor$ and treat them as categorical features, which is conducted by winning team of Criteo competition.

\textbf{3. iPinYou.}\footnote{http://contest.ipinyou.com/} iPinYou dataset is from iPinYou Global RTB(Real-Time Bidding) Bidding Algorithm Competition in 2013. We follow the data processing steps of~\cite{zhang2014real} and consider all 16 categorical features.

For all the datasets, we randomly split the examples into three parts: 70\% is for training, 10\% is for validation, and 20\% is for testing. We also remove each categorical features' infrequent levels appearing less than 20 times to reduce sparsity issue. Note that we want to compare the effectiveness and efficiency on learning higher-order feature interactions automatically, so we do not do any feature engineering but only feature transformation, e.g., numerical feature bucketing and categorical feature frequency thresholding.

\subsubsection{Evaluation Metrics}
We consider AUC and LogLoss for evaluating the performance of the models.

\textbf{LogLoss} LogLoss is both our loss function and evaluation metric. It measures the average distance between predicted probability and true label of all the examples.

\textbf{AUC} Area Under the ROC Curve (AUC) measures the probability that a randomly chosen positive example ranked higher by the model than a randomly chosen negative example. AUC only considers the relative order between positive and negative examples. A higher AUC indicates better ranking performance.

\subsubsection{Competing Models}
We compare xDeepInt with following models: LR(logistic regression)~\cite{mcmahan2011follow,mcmahan2013ad}, FM(factorization machine)~\cite{rendle2010factorization}, DNN (plain multilayer perceptron), Wide \& Deep~\cite{cheng2016wide}, DeepCrossing~\cite{shan2016deep}, DCN (Deep \& Cross Network)~\cite{wang2017deep}, PNN (with both inner product layer and outer product layer)~\cite{qu2016product,qu2018product}, DeepFM~\cite{guo2017deepfm}, xDeepFM~\cite{lian2018xdeepfm}, AutoInt~\cite{song2018autoint} and FiBiNET~\cite{huang2019fibinet}. Some of the models are state-of-the-art models for CTR prediction problem and are widely used in the industry.

\subsubsection{Reproducibility}
We implement all the models using Tensorflow~\cite{abadi2016tensorflow}. The mini-batch size is 4096, and the embedding dimension is 16 for all the features. For optimization, we employ Adam~\cite{kingma2014adam} with learning rate set to 0.001 for all the neural network models, and we apply FTRL~\cite{mcmahan2011follow,mcmahan2013ad} with learning rate as 0.01 for both LR and FM. For regularization, we choose L2 regularization with $\lambda$ = 0.0001 for dense layer. Grid-search for each competing model's hyper-parameters is conducted on the validation dataset. The number of DNN, Cross, CIN, Interacting layers is from 1 to 4. The number of neurons ranges from 128 to 1024. All the models are trained with early stopping and are evaluated every 2000 training steps.

For the hyper-parameters search of xDeepInt, The number of recursive feature interaction layers is from 1 to 4. For the number of sub-spaces $h$, the searched values are 1, 2, 4, 8 and 16. Since our embedding size is 16, this range covers from complete vector-wise interaction to complete bit-wise interaction. We use G-FTRL optimizer for embedding table and FTRL for PIN layers with learning rate as 0.01.

\subsection{Model Performance Comparison (Q1)}

\begin{table}[H]
	\caption{Performance Comparison of Different Algorithms on Criteo, Avazu and iPinYou Dataset.}\label{tbl1}
	\centering
	\resizebox{1.0\linewidth}{!}
        {\begin{tabular}{ccccccc}
			\hline
			& \multicolumn{2}{c}{Criteo}        & \multicolumn{2}{c}{Avazu}         & \multicolumn{2}{c}{iPinYou}         \\
			Model        & AUC             & LogLoss         & AUC             & LogLoss         & AUC             & LogLoss           \\ \hline
			LR           & 0.7924          & 0.4577          & 0.7533          & 0.3952          & 0.7692          & 0.005605          \\
			FM           & 0.8030          & 0.4487          & 0.7652          & 0.3889          & 0.7737          & 0.005576          \\
			DNN          & 0.8051          & 0.4461          & 0.7627          & 0.3895          & 0.7732          & 0.005749          \\
			Wide\&Deep   & 0.8062          & 0.4451          & 0.7637          & 0.3889          & 0.7763          & 0.005589          \\
			DeepFM       & 0.8069          & 0.4445          & 0.7665          & 0.3879          & 0.7749          & 0.005609          \\
			DeepCrossing & 0.8068          & 0.4456          & 0.7628          & 0.3891          & 0.7706          & 0.005657          \\
			DCN          & 0.8056          & 0.4457          & 0.7661          & 0.3880          & 0.7758          & 0.005682          \\
			PNN          & 0.8083          & 0.4433          & 0.7663          & 0.3882          & 0.7783          & 0.005584          \\
			xDeepFM      & 0.8077          & 0.4439          & 0.7668          & 0.3878          & 0.7772          & 0.005664          \\
			AutoInt      & 0.8053          & 0.4462          & 0.7650          & 0.3883          & 0.7732          & 0.005758          \\
			FiBiNET      & 0.8082          & 0.4439          & 0.7652          & 0.3886          & 0.7756          & 0.005679          \\
			xDeepInt     & \textbf{0.8111} & \textbf{0.4408} & \textbf{0.7675} & \textbf{0.3872} & \textbf{0.7791} & \textbf{0.005565} \\ \hline
	\end{tabular}}
\end{table}

The overall performance of different models is listed in \Cref{tbl1}. We have the following observations in terms of model effectiveness:
\begin{itemize}
	\item LR is generally worse than other algorithms, which indicates that learning higher-order feature interactions is essential for CTR model performance.
	\item FM brings the most significant boost in performance while we increase model complexity. This reveals the importance of learning explicit vector-wise feature interactions.
	\item Models that combining vector-wise and bit-wise interactions together consistently outperform other models. This phenomenon indicates that both types of feature interactions are essential to prediction performance and compensate each other.
	\item xDeepInt achieves the best prediction performance among all models. However, different datasets favor feature interactions of different degrees and bit-wise feature interactions of different complexity. The superior performance of our model could attribute to the fact that xDeepInt model the bounded degree of polynomial feature interactions by adjusting the depth of PIN and achieve different complexity of bit-wise feature interactions by changing the number of sub-spaces.
\end{itemize}

\subsection{Hyper-Parameter Study (Q2)}
In order to have deeper insights of the proposed model, we conduct experiments on three datasets and compare several variants of xDeepInt on different hyper-parameter settings.

\subsubsection{Depth of Network}
The depth of PIN determines the order of feature interactions. \Cref{tbl2} illustrates the performance change with respect to the number of PIN layers. When the number of layers set to 0, our model is equivalent to logistic regression and no interactions are learned. The performance of xDeepInt achieves the best when the number of layers is about 3 or 4. In this experiment, we set the number of sub-spaces as 1, to disable the bit-wise feature interactions.

\begin{table}[H]
	\caption{Impact of hyper-parameters: number of layers}\label{tbl2}
	\footnotesize
	\centering
	\resizebox{\linewidth}{!}
	{\begin{tabular}{c|ccccccc}
			\hline
			&\#Layers & 0      & 1      & 2      & 3      & 4      & 5      \\ \hline
			\multirow{2}{*}{Criteo}
			&AUC & 0.7921 & 0.8038 & 0.8050 & 0.8057 & \textbf{0.8063} & 0.8061 \\
			&LogLoss  & 0.4580 & 0.4477 & 0.4466 & 0.4461 & \textbf{0.4452} & 0.4454\\
			\hline
			\multirow{2}{*}{Avazu}
			&AUC & 0.7536 & 0.7654 & 0.7664 & \textbf{0.7675} & 0.7670 & 0.7662 \\
			&LogLoss  & 0.3951 & 0.3888 & 0.3879 & \textbf{0.3872} & 0.3875 & 0.3883 \\
			\hline
			\multirow{2}{*}{iPinYou}
			&AUC & 0.7690 & 0.7740 & 0.7775 & \textbf{0.7791} & 0.7783 & 0.7772 \\
			&LogLoss  & 0.005604 & 0.005576 & 0.005569 & \textbf{0.005565} & 0.005580 & 0.005571 \\
			\hline
	\end{tabular}}
\end{table}

\subsubsection{Number of Sub-spaces}
The subspace-crossing mechanism enables the proposed model to control the complexity of bit-wise interactions. \Cref{tbl3} demonstrates that subspace-crossing mechanism boosts the performance. In this experiment, we set the number of PIN layers as 3, which is generally a good choice but not the best setting for each dataset.

\begin{table}[H]
	\caption{Impact of hyper-parameters: number of sub-spaces}\label{tbl3}
	\footnotesize
	\centering
	\resizebox{\linewidth}{!}
	{\begin{tabular}{c|cccccc}
			\hline
			&\#Sub-spaces & 1      & 2      & 4      & 8      & 16     \\ \hline
			\multirow{2}{*}{Criteo}
			&AUC          & 0.8072 & 0.8081 & 0.8089 & 0.8096 & \textbf{0.8101} \\
			&LogLoss      & 0.4445 & 0.4435 & 0.4425 & 0.4421 & \textbf{0.4418} \\
			\multirow{2}{*}{Avazu}
			&AUC          & 0.7660 & 0.7668 & \textbf{0.7674} & 0.7672 & 0.7668 \\
			&LogLoss      & 0.3880 & 0.3877 & \textbf{0.3875} & 0.3878 & 0.3879 \\
			\multirow{2}{*}{iPinYou}
			&AUC          & 0.7772 & 0.7783 & \textbf{0.7788} & 0.7787 & 0.7784 \\
			&LogLoss      & 0.005590 & 0.005583 & \textbf{0.005568} & 0.005572 & 0.005580 \\
			\hline
	\end{tabular}}
\end{table}

\subsubsection{Activation Function}
By default, we use linear activation function on neurons of PIN layers. We also would like to explore how different activation function of PIN affect the performance. \Cref{tbl4} shows that the linear activation function is the most performant one for the PIN. We study the effect of activation function on Criteo dataset.

\begin{table}[H]
	\caption{Impact of hyper-parameters: activation function}\label{tbl4}
	\centering
	\begin{tabular}{c|cc}
		\hline
		& AUC     & LogLoss \\ \hline
		linear      & \textbf{0.8111}  & \textbf{0.4408}  \\
		tanh        & 0.8100  & 0.4418  \\
		sigmoid     & 0.8082  & 0.4434  \\
		softplus    & 0.8080  & 0.4436  \\
		swish       & 0.8100  & 0.4418  \\
		relu        & 0.8098  & 0.4419  \\
		leaky relu  & 0.8102  & 0.4415  \\
		elu         & 0.8099  & 0.4418  \\
		selu        & 0.8100  & 0.4418  \\ \hline
	\end{tabular}
\end{table}

\subsubsection{Optimizer}
We also build our model with Adam optimizer, same as all the competing models, to compare with our G-FTRL and FTRL combined optimization strategy. \Cref{tbl5} shows that our G-FTRL and FTRL combined optimization strategy achieves better performance. \Cref{tbl6} shows that our optimization strategy gets higher degree of feature sparse ratio (ratio of all zero embedding vectors in embedding table) and sparse ratio (ratio of zero weights in PIN layers), which results in lightweight model. One thing should be noted is that xDeepInt still achieves the best prediction performance among all models when using Adam optimizer, which demonstrates the effectiveness of xDeepInt architecture.

\begin{table}[H]
	\caption{Impact of hyper-parameters: optimizer}\label{tbl5}
	\centering
	\begin{tabular}{llll}
		\hline
		Dataset                  & Model     & LogLoss   & AUC    \\ \hline
		\multirow{2}{*}{Criteo}  & G-FTRL/FTRL  & \textbf{0.4408}    & \textbf{0.8111} \\
		& Adam & 0.4415    & 0.8105 \\ \hline
		\multirow{2}{*}{Avazu}   & G-FTRL/FTRL  & \textbf{0.3872}    & \textbf{0.7675} \\
		& Adam & 0.3873    & 0.7674 \\ \hline
		\multirow{2}{*}{iPinYou} & G-FTRL/FTRL  & \textbf{0.005565}  & \textbf{0.7791} \\
		& Adam & 0.005583  & 0.7784 \\ \hline
	\end{tabular}
\end{table}

\begin{table}[H]
	\caption{Analysis of model sparsity}\label{tbl6}
	\centering
	\begin{tabular}{c|cc}
		\hline
		Dataset     & Feature Sparse ratio     & Weight Sparse ratio \\ \hline
		Criteo      & 0.6506  & 0.1030  \\
		Avazu       & 0.2193  & 0.0448  \\
		iPinYou     & 0.8274  & 0.0627  \\ \hline
	\end{tabular}
\end{table}

\subsection{Ablation Study: Integrating Implicit Interactions (Q3)}
In this section, we conduct ablation study comparing the performance of our proposed model with and without integrating implicit feature interactions.

Feed-forward neural network is widely used by various model architectures for learning implicit feature interactions. In this experiment, we jointly train xDeepInt with a three-layer feed-forward neural network and name the combined model as xDeepInt+ to compare with vanilla xDeepInt.

\Cref{tbl7} compares vanilla xDeepInt and xDeepInt+. We observe that the jointly-trained feed-forward neural network does not boost the performance of vanilla xDeepInt. The reason is that vanilla xDeepInt model has already learned bit-wise interactions through the subspace-crossing mechanism. Thus, feed-forward neural network does not bring in additional predictive power.

\begin{table}[H]
	\caption{Ablation study comparing the performance of xDeepInt with and without integrating DNN}\label{tbl7}
	\centering
	\begin{tabular}{llll}
		\hline
		Dataset                  & Model     & LogLoss   & AUC    \\ \hline
		\multirow{2}{*}{Criteo}  & xDeepInt  & \textbf{0.4408}    & \textbf{0.8111} \\
		& xDeepInt+ & 0.4412    & 0.8107 \\ \hline
		\multirow{2}{*}{Avazu}   & xDeepInt  & \textbf{0.3872}    & \textbf{0.7675} \\
		& xDeepInt+ & 0.3874    & 0.7673 \\ \hline
		\multirow{2}{*}{iPinYou} & xDeepInt  & \textbf{0.005565}  & \textbf{0.7791} \\
		& xDeepInt+ & 0.005581  & 0.7787 \\ \hline
	\end{tabular}
\end{table}

\section{Conclusion}
In this paper, we design a novel network layer named polynomial interaction network (PIN), which learns the higher order vector-wise feature interactions on the embedding space. By incorporating PIN with the subspace-crossing mechanism, our proposed model xDeepInt learns bit-wise and vector-wise feature interactions of bounded degree simultaneously in controllable manner. We add residual connections to PIN layers, such that the output of each layer is an ensemble of the low-order and high-order interactions. The degree of interaction is controlled by the number of PIN layers, and the complexity of bit-wise interaction is controlled by the number of sub-spaces. Additionally, an optimization method is introduced to performs feature selection and interaction selection based on the network structure. Our experimental result demonstrates that the proposed xDeepInt outperforms existing state-of-art methods on real-world datasets. To our best knowledge, xDeepInt is the first neural network architecture that achieves state-of-art performance without integration of feed-forward neural network using non-linear activation functions.

We have multiple directions of future work. First, the proposed model only focuses on modeling fixed-length feature vectors. In order to model historical and sequential behavior in recommendation systems\cite{zhou2018deep,zhou2018deepevo}, We are interested in making our model architecture applicable to variable-length feature vectors. Second, We would like to extend the application of polynomial interaction layers to more modeling scenarios and exploit PIN's potential on other problems. Third, the model is fully explainable when the subspace crossing mechanism is disable. The explainability of the model is another direction of future work.

\newpage
\bibliographystyle{ACM-Reference-Format}
\bibliography{xdeepint-ref.bib}
\nocite{*}

\end{document}